\documentclass{article}

\usepackage{url}

\usepackage[english]{babel}

\usepackage[utf8]{inputenc}
\usepackage{xspace}
\usepackage{amsmath,amsthm,amssymb,mathtools}
\usepackage{lmodern}

\usepackage{algorithm}
\usepackage[noend]{algorithmic}

\usepackage{rotating}

\newcommand{\MMAS}{MMAS\xspace}

\usepackage{tikz}
\usepgflibrary{arrows}
\usetikzlibrary{arrows}

\usepackage[algo2e,ruled,vlined,linesnumbered]{algorithm2e}

\usepackage{xcolor}
\usepackage{tikz}
\usepackage{graphicx}

\allowdisplaybreaks[4]
\newtheorem{theorem}{Theorem}
\newtheorem{lemma}[theorem]{Lemma}

\newtheorem{definition}[theorem]{Definition}

\newcommand{\oea}{$(1 + 1)$~EA\xspace}

\newcommand{\oplea}{$(1+\lambda)$~EA\xspace}
\newcommand{\mpoea}{$(\mu+1)$~EA\xspace}

\newcommand{\om}{\textsc{OneMax}\xspace}

\newcommand{\R}{\ensuremath{\mathbb{R}}}

\newcommand{\N}{\ensuremath{\mathbb{N}}}

\newcommand{\ignore}[1]{}

\usepackage{multirow}
\usepackage{booktabs}                                     % pretty tables

\newcommand*{\nwspace}{\hspace*{.1em}}
\newcommand*{\Or}{\mathrm O}
\newcommand*{\xorg}{x_{\mathrm{org}}}
\newcommand*{\dominates}{\succcurlyeq}
\newcommand*{\sdominates}{\succ}
\newcommand*{\mutate}{\textup{mutate}}
\newcommand*{\cross}{\textup{cross}}
\newcommand*{\uar}{u.a.r.\xspace}
\newcommand*{\wrt}{w.r.t.\xspace}
\newcommand{\ie}{\hbox{i.\hspace{0.125em} e.}\xspace}                  % i.e.

\newcommand{\mplea}{$(\mu+\lambda)$~EA\xspace}
\newcommand{\RLS}{RLS\xspace}

\newcommand{\etal}{et al.}
\let\oldnl\nl												% Store \nl in \oldnl
\newcommand*{\nonl}{\renewcommand{\nl}{\let\nl\oldnl}}		% Remove line number for one line

\begin{document}
	
	\title{Analysis of Evolutionary Algorithms in Dynamic and Stochastic Environments}
	\author{Vahid Roostapour
 \and Mojgan Pourhassan \and Frank Neumann \and \\ Optimisation and Logistics\\ School of Computer Science\\ The University of Adelaide\\Australia}
%\institute{Optimisation and Logistics, School of Computer Science, The University of Adelaide}
	\maketitle

%	\begin{abstract}

%
%		List of papers:
%		
%		Droste dynamic \om~\cite{DrosteDynamic, DrosteDynamic2003}
%		
%		Same results of Droste with easier analysis in addition to some new results~\cite{WittDynamic}
%		
%		%Witt new Journal paper~\cite{LissovoiWittDynamic2017}
%		
%		Witt MMAS Versus Population-Based EA MM~\cite{LissovoiWittDynamic2016}
%		
%		%Witt I don't even understand the title~\cite{LissovoiWittMigrationTopology2016}
%		
%		Witt ACO dynamic shortest path problem~\cite{LissovoiWittDynamic2015}
%		
%		Our Dynamic VCP~\cite{UsDVCGecco2015} and the new one FOCI
%		
%		Dynamic makespan scheduling~\cite{neumann2015runtime}
%		
%		Feng Linear Functions and Dynamic Uniform Constraints~\cite{FengLinearFunctionDynamicConstraintsGecco2017}
%		
%		Timo stochastic \om~\cite{giessen2016robustness}
%		
%		Jansen journal paper\cite{JansenDynamic}
%		
%		Bennett large noise variance~\cite{BennettFoga2015}
%		
%		Timo and Tobias recombination and noise~\cite{TobiasRecombinationNoise2015}
%		
%		Timo and Tobias ACO noisy \om~\cite{TobiasACONoise2016}
%		
%		Timo and Tobias robustness under small perturbation~\cite{TobiasRobust2016}
%		
%		Timo and Tobias Distribution study for noise~\cite{friedrich2016graceful}
%		
%		Timo and Tobias 2016~\cite{TobiasNoise2016}
%		
%		Timo and Tobias 2017~\cite{TobiasTimoNoise2017}
%		
%	\end{abstract}
\begin{abstract}
	Many real-world optimization problems occur in environments that change dynamically or involve stochastic components. Evolutionary algorithms and other bio-inspired algorithms have been widely applied to dynamic and stochastic problems.  This survey gives an overview of major theoretical developments in the area of runtime analysis for these problems. We review recent theoretical studies of evolutionary algorithms and ant colony optimization for problems where the objective functions or the constraints change over time. Furthermore, we consider stochastic problems under various noise models and point out some directions for future research.
\end{abstract}

\newpage
		\tableofcontents
		\newpage
	\section{Introduction}

Real-world problems are often stochastic and have dynamic components. Evolutionary algorithms and other bio-inspired algorithmic approaches such as ant colony optimization have been applied to a wide range of stochastic and dynamic problems. The goal of this chapter is to give an overview on recent theoretical developments in the area of evolutionary computation for stochastic and dynamic problems in the context of discrete optimization.

Stochastic problems occur frequently in real-world applications due to unpredictable factors. A good example is the scheduling of trains. Schedules give precise timing when trains arrive and depart. However, the actual departure and arrival times may be subject to delays due to various factors such as weather conditions and interfering schedules of other trains. Using evolutionary computation for the optimization of stochastic problems, the uncertainty  is usually reflected through a noisy fitness function.
The underlying fitness function for these problems is noisy in the sense that it produces different results for the same input. Two major noise models, namely prior noise and posterior noise, have been introduced and investigated in the literature. In the case of prior noise, the solution is changed prior to the evaluation of the given fitness function, whereas in the case of posterior noise the solution is evaluated with the given fitness function and a value according to a given noise distribution is added before returning the fitness value. 

Dynamic problems constitute another important part occurring in real-world applications. Problems can change over time due to different components becoming unavailable or available at a later point in time. Different parts of the problem that can be subject to a change are the objective function and possible constraints of the given problem. In terms of scheduling of trains, trains might become unavailable due to mechanical failures and it might be necessary to reschedule the trains in the network in order to still serve the demands of the customers well.

The area of runtime analysis has contributed many interesting studies to the theoretical understanding of bio-inspired algorithms in this area. 
We start by investigating popular benchmark algorithms such as randomized local search (RLS) and \oea on different dynamic problems. This includes dynamic versions of \om, the classical vertex cover problem, the makespan scheduling problem, and problem classes of the well-known knapsack problem. Afterwards, we summarize main results for stochastic problems. Important studies in this area consider the \om problem and investigate the runtime behavior of evolutionary algorithms with respect to prior and posterior noise. Moreover, the influence of populations in evolutionary algorithms for solving stochastic problems is analyzed in the literature, and we place a particular emphasis on those studies.  Furthermore, we review the performance of the population based algorithms on different posterior noise functions. 

Ant colony optimization (ACO) algorithms are another important type of bio-inspired algorithms that has been used and analyzed for solving dynamic and stochastic problems. Due to their different way of constructing solutions, based on sampling from the underlying search space by performing random walks on a so-called construction graph, they have a different ability to deal with dynamic and stochastic problems. Furthermore, an important parameter in ACO algorithms is the pheromone update strength which allows to determine how quickly previously good solutions are forgotten by the algorithms. This parameter plays a crucial role when distinguishing ACO algorithms from classical evolutionary algorithms.
 At the end of this chapter, we present a summary of the obtained results on the dynamic and stochastic problems in the context of ACO.

 This chapter is organized as follows. In Section~\ref{sec:prelim}, we summarize the dynamic and stochastic settings that have been investigated in the literature. We present the main results obtain for evolutionary algorithms in  dynamic environments and stochastic environments in Section~\ref{sec:dyn} and~\ref{dynevsec:noise}, respectively. We highlight theoretical results on the behavior of ACO algorithms for dynamic and stochastic problems in Section~\ref{sec:ACO}. Finally, we finish with some conclusions and outline some future research directions.

	\section{Preliminaries}
	\label{sec:prelim}
This section includes the formal definitions of dynamic and stochastic optimization settings that are investigated in this chapter. 

%Moreover, randomized local search and evolutionary algorithms that are frequently investigated in %theoretical studies are presented in this section.
In dynamically changing optimization problems, some part of the problem is subject to change over time. Usually changes to the objective function or the constraints of the given problem are considered. The different problems that have been studied from a theoretical perspective will be introduced in the forthcoming subsections.
 %Dynamic changes on the fitness function have been addressed in~\cite{Shift functions} in which a shift of the fitness function on the search space is considered. In this chapter, we focus on the problems with dynamically changing constrains and the problems in which small changes on the instance of the problem are made. 
In the case of stochastic optimization problems, the optimization algorithm does not have access to the deterministic fitness value of a candidate solution.
Different types of noise that change the actual fitness value have been introduced. The most important ones, prior noise and posterior noise, will be introduced in Section~\ref{sec:defnoise}.
	
The theoretical analysis of evolutionary algorithms for dynamic and stochastic problems concentrates on the classical algorithms such as randomized local search (\RLS) and \oea. Furthermore, the benefit of population-based approaches has been examined. These algorithms will be introduced in Section~\ref{sec:algos}.

\subsection{Dynamic \om Problem}
	\label{dynevsec:prel:DynamicOneMax}
Investigations have started by considering a generalization of the the classical \om problem.
	In the \om problem, the number of ones in the solution is the objective to be maximized. Droste~\cite{DrosteDynamic} has interpreted this problem as maximizing the number of bits that match a given objective bit-string. Based on this, he has introduced the dynamic \om problem, in which dynamic changes happen on the objective bit-string over time. An extended version of this problem is defined by K\"{o}tzing \etal~\cite{WittDynamic} where not only bit-strings are allowed. Here each position can take on integer values in $\{0, \ldots, r-1\}$ for $r\in \N_{\geq 2}$. The formal definition of the problem follows.
	
	Let $[r]= \{0, \ldots, r-1\}$ for $r\in \N_{\geq 2}$, and $x, y \in [r]$. Moreover, let the distance between $x$ and $y$ be 
$$d(x,y)= \min\left\{(x-y)\mod r,(y-x) \mod r \right\}.$$ 

The extended \om problem, $\text{\om}_a:[r]^n \rightarrow \R$, where $a$ is the objective string defining the optimum, is given as:
	$$\text{\om}_a(x) = \sum_{i=1}^{n} d(a_i,x_i).$$
	The goal is to find and maintain a solution with minimum value of $\text{\om}_a$. 
	
Given a probability value $p$, the dynamism that is defined on this problem is to change each component $i$, $1 \leq i \leq  n$,
of the optimal solution $a$ independently as:

    \begin{table}[ht]
\begin{center}
    \begin{tabular}{c c}
    \multirow{3}{*}{$a_i= \Biggl\{ $} 
    & $ a_i+1 \mod r;$ \ \ \ with probability $p/2$ \\
    & $ a_i-1 \mod r;$ \ \ \ with probability $p/2$ \\
    & $ a_i;$ \ \ \ \ \ \ \ \ \ \ \ \ \ \ \ \   with probability $1-p$
    \end{tabular}
\end{center}
\end{table}

	\subsection{Linear Pseudo-Boolean Functions Under Dynamic Uniform Constraints}
	\label{dynevsec:prel:DynamicConstraints}
Linear pseudo-Boolean functions play a key role in the runtime analysis of evolutionary algorithms.
	Let $x=x_1x_2 \ldots x_n$ be a search point in search space $\{0,1\}^n$, and $w_i$,  $1 \leq  i \leq n$  positive real weights. A linear pseudo-Boolean function $f(x)$ is defined as:
	$$f(x)= w_0 + \sum^{n}_{i=1} w_ix_i.$$

For simplicity and as done in most studies, we assume $w_0=0$ in the following.
	The optimization of a linear objective function under a linear constraint is equivalent to the classical knapsack problem~\cite{KnapsackProblems}. 
The optimization of a linear objective function together with a uniform constraint has recently been investigated in the static setting~\cite{Tobias2017UniformConstraints}. Given a bound $B$, $0\leq B\leq n$, a solution $x$ is feasible if the  number of $1$-bits of the search point $x$ is at most $B$. The bound $B$ is also known as the \textit{cardinality bound}.
	We denote the number of $1$-bits of $x$ by $|x|_1 = \sum_{i=1}^n x_i$. The formal definition for maximizing a pseudo-Boolean linear function under a cardinality bound constraint is given by:
	\begin{eqnarray*}
		\ \ \ \max f(x)\\
		\text{s.t.} \ |x|_1 \leq B.
	\end{eqnarray*}
	
	The dynamic version of this problem, referred to as  the problem with a \emph{dynamic uniform constraint}, is defined in~\cite{FengLinearFunctionDynamicConstraintsGecco2017}. Here the cardinality bound changes from $B$ to some new value $B^*$. Starting from a solution  that is optimal for the bound $B$, the problem is then to find an optimal solution for $B^*$. The re-optimization time of an evolutionary algorithm is defined as the number of fitness evaluations that is required to find the new optimal solution. 
	
\subsection{Dynamic Vertex Cover Problem}
\label{dynevsec:prel:DynamicVertexCover1}
The vertex cover problem is one of the best-known NP-hard combinatorial optimization problems.
Given a graph $G=(V,E)$, where $V=\{v_1, \ldots, v_n\}$ is the set of vertices and $E=\{e_1, \ldots, e_m\}$ is the set of edges, the goal is to find a minimum subset of nodes $V_C \subseteq V$ that covers all edges in $E$, i.e. $\forall e \in E , e\cap V_C \ne \emptyset$.
In the dynamic version of the problem, an edge can be added to or deleted from the graph.

As the vertex cover problem is NP-hard, it has been mainly studied in terms of approximations. The problem can be approximated within a worst case approximation ratio of $2$ by various algorithms. One standard approach to obtain a $2$-approximation is to compute a maximal matching and take all nodes adjacent to the chosen matching edges for the vertex cover.
Starting from a solution that is a 2-approximation for the current instance of the problem, in the dynamic version of the problem the goal is to obtain a $2$-approximate solution for that instance of the problem after one dynamic change. The re-optimization time for this problem refers to the required time for the investigated algorithm to find a 2-approximate solution for the new instance. This dynamic setting has been investigated in~\cite{UsDVCGecco2015}. %,DynamicVertexCoverFOCI2017

\subsection{Dynamic Makespan Scheduling Problem}

The makespan scheduling problem can be defined as follows. Given $n$ jobs and their processing times $p_i>0$, $1\leq i \leq n$, the goal is to assign each job to one of two machines $M_1$ and $M_2$ such that the makespan is minimized. The makespan is the time that the busier machine takes to finish all assigned jobs. A solution is represented by a vector $x\in \{0,1\}^n$ which means that job $i$ is assigned to machine $M_1$ if $x_i=0$ and it is assigned to $M_2$ if $x_i = 1$, $1\leq i\leq n$. With this representation, the makespan of a given solution $x$ is given by 
$$f(x)=\max\left\{\sum_{i=1}^{n}{p_i(1-x_i)},\sum_{i=1}^{n}{p_ix_i}\right\}$$ and the goal is to minimize $f$. In the dynamic version of this problem,
the processing time of a job may change over time, but stays within a given interval. In \cite{neumann2015runtime}, the setting $p_i \in [L, U]$, $1 \leq i \leq n$, where $L$ and $U$ are a lower and upper bound on each processing time, have been investigated. The analysis concentrates on the time evolutionary algorithms need to produce a solution where the two machines have discrepancy at most $U$.
 Dynamic changes to the  processing times of the jobs have been investigated in two different settings. In the first setting, an adversary is allowed to change the processing time of exactly one job. In the second setting, the job to be changed is picked by an adversary but the processing time of a job is altered randomly.
% an adversary may change the processing time of a job. The optimization times of \oea and \RLS have %been considered against the dynamic makespan scheduling in two cases: \textbf{either the adversary is %able to make a change in each point of time, or the re-optimization time which the algorithm takes to %optimize a changed optimal solution.   }

%\merk{Hi Vahid! It is not clear what you mean here!}

	\subsection{Stochastic Problems and Noise Models}
\label{sec:defnoise}
We consider stochastic optimization problems where the fitness function is subject to some noise.
	Two different noise models have mainly been studied in the area of the theoretical analysis of evolutionary computation. Noises that affect the solution before the evaluation are called \emph{prior noise}. In this case, the fitness function returns the fitness value of a solution that may differ from the given solution because of the noise. Droste studied the effect of a prior noise which flips one randomly chosen bit of the given solution with probability of $p$, before each evaluation \cite{droste2004analysis}. Note that the noise does not change the solution, but it causes the fitness function to evaluate a solution with a noisy bit flip. Other kinds of prior noises have also been considered. For example, a prior noise which flips each bits with the probability of $p$ or a prior noise which sets each bit independently with probability of $p$ to 0 \cite{giessen2016robustness}.% \merk{Check this!} Corollary 9.
	
Another important type of noise is where the fitness of the solution is changed after evaluation. This type of noise is called \emph{posterior noise} or \emph{additive posterior noise}. The noise which commonly comes from a defined distribution $D$, adds the value of a random variable sampled from $D$ to the value coming from the original fitness function~\cite{TobiasTimoNoise2017,friedrich2016graceful,giessen2016robustness}.
	
	In the noisy environment, the problem of finding the optimal solutions is harder as the noise is misleading the search. The goal is to find an optimal solution for the original non noisy fitness function by evaluating solutions on the fitness function affected by noise. However, it has been proven that simple evolutionary algorithms behave considerably well when facing this kind of problems. In addition to this, properties of stochastic settings that are hard for evolutionary algorithms to deal with have also been studied in \cite{friedrich2016graceful}. 

We concentrate on stochastic problems with a fixed and known solution length that are subject to noise. We would like to mention that there are also studies investigating the performance of evolutionary algorithms with unknown solution length. This poses a different type of uncertainty which we will not capture in this chapter. We refer the interesting reader to \cite{DBLP:conf/foga/CathabardLY11,DBLP:conf/gecco/DoerrDK15}.

\subsection{Evolutionary algorithms}
\label{sec:algos}
Analyzing evolutionary algorithms often starts by investigating a standard randomized local search approach and a simple \oea. Here we present these algorithms in addition to a population-based \mplea for which results are summarized in Section~\ref{dynevsec:noise}.

A standard \RLS (see Algorithm~\ref{dynevalg:RLS}) starts with a bit-string as the initial solution, makes a new solution by flipping one bit of the current solution uniformly at random at each iteration, and replaces the current solution with the new solution if the new one is better in terms of fitness. The algorithm repeats these steps, while the stopping criterion is not met.

	\begin{algorithm2e}[t]
		\caption{\RLS}
		\label{dynevalg:RLS}
		\SetAlgoSkip{tinyskip}
		The initial solution $x$ is given\;
		\While{stopping criterion not met}{
			$y$ $\gets$ flip one bit of $x$ chosen uniformly at random\;
			\If {$f(y) \geq f(x)$}{$x \gets y$\;}
		}
	\end{algorithm2e}

The \oea (see Algorithm~\ref{dynevalg:oneplusone}) is a simple evolutionary algorithm in which the population consists of only one solution and only one solution is generated at each time step. This algorithm is quite similar to \RLS, except that multiple bit flips are allowed at each iteration. Instead of flipping one bit uniformly at random, in this algorithm all bits of the current solution are flipped with probability $1/n$, where $n$ is the size of the solution.

	\begin{algorithm2e}[t]
		\caption{\oea}
		\label{dynevalg:oneplusone}
		\SetAlgoSkip{tinyskip}
		The initial solution $x$ is given\;
		\While{stopping criterion not met}{
			$y$ $\gets$ flip each bit of $x$ independently with probability $1/n$\;
			\If {$f(y) \geq f(x)$}{$x \gets y$\;}
		}
	\end{algorithm2e}
A classical question in the area of evolutionary computation is whether populations help to achieve better results compared to algorithms working at each time step with a single solution.
The \mplea (Algorithm~\ref{dynevalg:mupluslamda}) is the population based version of \oea. In this algorithm, $\mu$ denotes the size of the parent population. In each iteration, the algorithm creates $\lambda$ offspring by mutating $\lambda$ parents which have been chosen uniformly at random from the parent population. Finally all the solutions from parents and offspring are evaluated and $\mu$ best ones (in terms of fitness function) survive. They constitute the parent population of the next generation.
	\begin{algorithm2e}
	\caption{\mplea}
	\label{dynevalg:mupluslamda}
	\SetAlgoSkip{tinyskip}
	$P$ is a set of $\mu$ uniformly chosen solutions\;
	\While{stopping criterion not met}{
		$O\gets\emptyset$\;
		\For{$i=1$ \KwTo $\lambda$}{
			pick $x$ u.a.r. from $P$\;
			$y$ $\gets$ flip each bit of $x$ independently with probability $1/n$\;
			$O\gets O\cup y$\;
		}
		\For{$x\in P\cup O$}{\text{evaluate} $f(x)$\;}
		$P \gets \mu$ $f$-maximal elements from $P \cup O$\;
	}
\end{algorithm2e}

One of the questions raised by using the population is the effect of crossover operator in the robustness. This has been investigated by Friedrich~\etal in~\cite{friedrich2016robustness}. To this end, they considered a framework consists of one wide and many narrow parallel paths, with equal distances, for solutions to achieve the highest fitness value. Moreover, the fitness grows more quickly along narrow paths and a solution which is not located in one of the paths does not survive. It is shown that algorithms with higher recombination rate optimize through the wide path while narrow paths are more favored by algorithms with zero recombination rate. A change that moves the framework along $x$-axis will cause the extinction of solutions on the narrow paths, however, solutions on the wide path may survive. This shows the benefit of crossover operation for the robustness of the algorithms using population.

Analyzing evolutionary algorithms with respect to their runtime behavior, one considers the number of solutions that are produced until a solution of desired quality has been achieved. The expected time to reach this goal refers to the expected number of such solutions. The \emph{expected  optimization time} refers to the expected number of solutions that are produced until an optimal search point has been produced for the first time. Considering dynamic problems, we are often interested in the \emph{expected re-optimization time} of an algorithm. Starting with a good (or even optimal) solution for the considered problem, the expected number of constructed solutions required to obtain a solution of the same quality after a dynamic change has occurred is analyzed.

	\section{Analysis of Evolutionary Algorithms on Dynamic Problems}
\label{sec:dyn}
%	\subsection{Introduction}
In this section, we summarize recent theoretical analyses that have been performed on evolutionary algorithms dealing with dynamic optimization problems. In~\cite{FengLinearFunctionDynamicConstraintsGecco2017, Feng2018Algorithmica}, the efficiency of evolutionary algorithms for solving linear pseudo-Boolean functions with a dynamic linear constraint has been investigated.  Particular attention has been paid to the \om problem. \om has been the center of attention in some other related works as well~\cite{DrosteDynamic, WittDynamic}. We first present the investigations that have been performed on this problem, then we give a summary of the results that have been obtained for linear pseudo-boolean functions under dynamic uniform constraints. Furthermore, in this section we explain the analysis that has been carried out for the dynamic vertex cover problem and the makespan scheduling problem. Another problem which has been investigated in the context of dynamic optimization is the \textsc{Maze} problem for which evolutionary algorithms as well as ant colony optimization algorithms have been theoretically studied~\cite{DBLP:conf/ppsn/KotzingM12, Lissovoi2016, Lissovoi2017}. The results of evolutionary algorithms and ACO algorithms for this problem are presented in Section~\ref{sec:eamaze} and Section~\ref{sec:ACO}, respectively.
	
	\subsection{\om Under Dynamic Uniform Constraints}
	\label{dynevsec:OneMax}
	The first runtime analysis of evolutionary algorithms for a dynamic discrete problem has been presented by Droste~\cite{DrosteDynamic}. In that article, the \om problem is considered and the goal is to find a solution which has the minimum Hamming distance to an objective bit-string. A dynamic change in that work is changing one bit of the objective bit-string, which happens at each time step with probability $p'$ and results in the dynamic changes of the fitness function over time. Droste has found the maximum rate of the dynamic changes such that the expected optimization time of \oea remains polynomial for the studied problem. More precisely, he has proved that \oea has a polynomial expected runtime if $p'=O(\log(n)/n)$, while for every substantially larger probability the runtime becomes super polynomial. It is worth noting that the results of that article hold even if the expected re-optimization time of the problem is larger than the expected time until the next dynamic change happens. 
	
Using drift analysis, K\"{o}tzing \etal~\cite{WittDynamic} have reproved some of the results in~\cite{DrosteDynamic}. Furthermore, they have carried out theoretical investigations for the extended dynamic \om problem (see Section~\ref{dynevsec:prel:DynamicOneMax}), in which each variable can take on more than two values.  They also carried out an \textit{anytime analysis} (introduced in~\cite{Jansen2014BiStableDynamicOpt}) and show how closely their investigated algorithm can track the dynamically moving target over time.

%\textbf{i was planing to bring some details on this paper but I have not done it yet.}	

	The optimization time of evolutionary algorithms for \om and the general class of  linear pseudo-Boolean function, under a dynamic uniform constraint given in Section~\ref{dynevsec:prel:DynamicConstraints} has been analysed in~\cite{FengLinearFunctionDynamicConstraintsGecco2017, Feng2018Algorithmica}. For now, we concentrate on \om with with dynamic uniform constraint.
	The authors have analysed a standard \oea (Algorithm~\ref{dynevalg:oneplusone}) and three other evolutionary algorithms which are presented in Algorithms~\ref{dynevalg:MOEA} to~\ref{dynevalg:MOGA}. The results of their investigations are summarised in Table~\ref{dynevtable:OneMaxResults}. The \oea analysed in this paper, uses the fitness function
	$$f_{(1+1)}(x)= f(x) - (n +1) \cdot \max \nwspace \{0, |x|_1 - B^*\}$$
already introduced in~\cite{Tobias2017UniformConstraints}. It gives a large penalty to infeasible solutions by subtracting for each unit of constraint violation a term of $(n+1)$. This implies that each infeasible solution is worse than any feasible one.
	The penalty of this fitness function, guides the search towards the feasible region and does not allow the \oea to accept an infeasible solution after a feasible solution has been found for the first time.

	\begin{sidewaystable}
		\begin{center}
			\renewcommand{\arraystretch}{1.8}
			\begin{tabular}{@{}lccccc@{}}
				\toprule
				\textbf{\oea} & \textbf{MOEA} & \textbf{MOEA-S} & \textbf{MOGA} &\\
				\midrule
				$\Or \! \left(n \nwspace \log\!\left(\frac{n-B\ \,}{n- B^*}\!\right) \!\right)$ &
				$\Or \! \left( nD \nwspace  \log\!\left( \frac{n-B\ \,}{n- B^*}\!\right) \!\right)$ &
				$\Or \! \left( n \nwspace  \log\!\left( \frac{n-B\ \,}{n- B^*}\!\right) \!\right)$ &
				$\Or \! \left(\min\{\sqrt{n} D^{\frac{3}{2}}, D^2 \sqrt{\frac{n}{n-B^*}}\}\!\right)$ &
				if $B < B^*$ \\
				$\Or \! \left(n \nwspace  \log \!\left( \frac{B\ \,}{B^*}\!\right) \!\right)$ &
				$\Or \! \left(nD \nwspace \log \!\left( \frac{B\ \,}{B^*}\!\right) \!\right)$ &
				$\Or \! \left( n \nwspace  \log \! \left( \frac{B\ \,}{B^*} \!\right) \!\right)$ &
				$\Or \! \left(\min\{\sqrt{n} D^{\frac{3}{2}}, D^2 \sqrt{\frac{n}{B^*}}\} \!\right)$ &
				if $B > B^*$\\
				\bottomrule\\
			\end{tabular}
		\end{center}
		\caption{Upper bounds on the expected re-optimization times of evolutionary algorithms on the \om problem with a dynamic uniform constraint.}
		\label{dynevtable:OneMaxResults}
	\end{sidewaystable}
	
	Shi~\etal~\cite{FengLinearFunctionDynamicConstraintsGecco2017, Feng2018Algorithmica} have used multiplicative drift analysis~\cite{algorithmica/DoerrJW12} for investigating the behavior of the studied algorithms. The potential function that they have used for analyzing \oea on \om with a dynamic uniform constraint is $|x|_0$, when $B\leq B^*$. Here, the initial solution, denoted by $x_{org}$, is feasible, and the algorithm needs to increase the number of ones of the solution, until the cardinality bound $B^*$ is reached. In this situation, the drift on $|x|_0$ is $\Omega(|x|_0/n)$ for \oea.
Using multiplicative drift analysis, the expected number of generations to reach a solution $x^*$ with $|x^*|_0=n-B^*$ is 
$$\Or \left(n \log \left(\frac{|\xorg|_0}{|x^*|_0}\right)\right)= \Or \! \left(n \nwspace \log\!\left(\frac{n-B\ \,}{n- B^*}\!\right) \!\right).$$ 
	
	For the situation where $B\geq B^*$, the initial solution is infeasible and the number of ones of the solution need to decrease (and possibly increase again, in case the last move to the feasible region has decreased $|x|_1$ to less than $B^*$). The considered potential function in this situation is $|x|_1$ and the drift on that is $\Omega(|x|_1/n)$, giving an expected re-optimization time of $\Or\left(n \nwspace \log\!\left(\frac{B}{B^*}\!\right) \!\right)$.

	\begin{algorithm2e}[t]
		\caption{MOEA; Assuming $B \le B^*\!\!.$~\cite{FengLinearFunctionDynamicConstraintsGecco2017}}
		\label{dynevalg:MOEA}
		\SetAlgoSkip{tinyskip}
		$P \gets$ an initial solution\;
		\While{stopping criterion not met}{
			Choose $x \in P$ uniformly at random\;
			Obtain $y$ from $x$ by flipping each bit of $x$ with probability $1/n$\;
			\If {$(B^* \nwspace {\ge} \nwspace |y|_1 \nwspace {\ge} \nwspace B) \wedge (\nexists \nwspace w \in P \colon w \dominates_{\textup{MOEA}} y)$}{
				$P \gets (P \cup \{y\}) \setminus \{z \in P \mid y \sdominates_{\textup{MOEA}} z \}$\;
			}
		}
	\end{algorithm2e}
	
	The second algorithm that the authors have investigated is the Multi-Objective Evolutionary Algorithm (MOEA) (see Algorithm~\ref{dynevalg:MOEA}). Here dominance of solutions is defined with respect to the vector-valued fitness function 
$$f_{\text{MOEA}}(x)=(|x|_1, f(x)).$$

 A solution $y$ dominates a solution $z$ \wrt $f_{\text{MOEA}}$ ($y \succeq z$) iff $|y|_1=|z|_1$ and $f(y)\geq f(z)$. Furthermore, $y$ strictly dominates $z$ ($y \succ z$) iff $y \succeq z$ and $f(y) > f(z)$. The algorithm keeps at most one individual for each Hamming weight between $B$ and $B^*$. Let $D=|B^*-B|$, then the size of population $P$ is at most $D+1$. The analysis shows that this population size slows down the re-optimization process for the \om problem. For the case where $B<B^*$ and $B>B^*$, the potential function that Shi~\etal~\cite{FengLinearFunctionDynamicConstraintsGecco2017} have used for analyzing this algorithm is $M=\min_{x\in P}|x|_0$ and $M=\max_{x\in P}|x|_1$, respectively. The analysis is similar to their analysis of \oea, except that the drift on $M$ is $\Omega(\frac{M}{n\cdot D})$. The $D$ in the denominator comes from the fact that selecting the individual $x$ with minimum $|x|_0$ for $B< B^*$ (minimum $|x|_1$ for $B>B^*$)  from the population, happens at each iteration with probability at least $\frac{1}{D+1}$. Using multiplicative drift analysis,
they obtained an upper bound of $\Or \! \left( nD \nwspace  \log\!\left( \frac{n-B\ \,}{n- B^*}\!\right) \!\right)$ for $B < B^*$ and an upper bound of $\Or \! \left(nD \nwspace \log \!\left( \frac{B\ \,}{B^*}\!\right) \!\right)$ for $B > B^*$.
	
	\begin{algorithm2e}
		\caption{MOEA-S; Assuming $B \le B^*\!\!.$~\cite{FengLinearFunctionDynamicConstraintsGecco2017}}
		\label{dynevalg:MOEA-S}
		\SetAlgoSkip{tinyskip}
		$P \gets$ an initial solution\;
		\While{stopping criterion not met}{
			Choose $x \in P$ uniformly at random\;
			Obtain $y$ from $x$ by flipping bit one bit $x_i$, $i \in \{1,\dots,n\}$ chosen \uar\;
			\If {$\forall \nwspace z \in P \colon y \parallel_{\mathrm{MOEA-S}} z$\label{dynevline:MOEA-S_if_incomparable}}{
				$P \gets P \cup \{y\}$}
			\If {$(B^* \ge |y|_1 \ge B) \wedge (\exists \nwspace z \in P \colon y \dominates_{\textup{MOEA-S}} z)$\label{dynevline:MOEA-S_if_dominating}}{
				$z \gets y $\;
			}
		}
	\end{algorithm2e}
	
	The third investigated algorithm is a variant of MOEA named MOEA-S shown in Algorithm~\ref{dynevalg:MOEA-S}. In this algorithm only single-bit flips are allowed and a different definition for dominance is used. The new notion of dominance does not let the population size grow to a size larger than 2. If $B \le B^*$, for two bit strings $y,z \in \{0,1\}^n$ we have:
	\begin{itemize}
		
		\item $y$ dominates $z$, denoted by $y \dominates_{\mathrm{MOEA-S}} z $ if \emph{at most one} value among $|y|_1$ and $|z|_1$ equals $B^*$ or $B^*-1$, and $(|y|_1 > |z|_1) \vee (|y|_1 = |z|_1 \wedge f(y) \ge f(z))$
		\item $y$ dominates $z$, denoted by $y \dominates_{\mathrm{MOEA-S}} z $ if both $|y|_1,|z|_1 \in \{B^*,B^*-1\}$, and $|y|_1 = |z|_1 \wedge f(y) \ge f(z)$

	\end{itemize}
This implies that $y$ and $z$ are incomparable, denoted by $y \parallel_{\mathrm{MOEA-S}} z$, iff $|y|_1 = B^*$ and $|z|_1 = B^*-1 $ or vice versa.

	For $B > B^*$, a similar definition of dominance is given by switching the dependency of $|y|_1 \ge |z|_1$ on the number of $1$-bits to $|y|_1 \le |z|_1$.
	The results of MOEA-S are obtained by observing that this algorithm behaves like \RLS on \om. It is shown that the expected re-optimization time for \om with a dynamic uniform constraint is $\Or \! \left( n \nwspace  \log\!\left( \frac{n-B\ \,}{n- B^*}\!\right) \!\right)$ if $B< B^*$ and $\Or \! \left( n \nwspace  \log \! \left( \frac{B\ \,}{B^*} \!\right) \!\right)$ if $B > B^*$.

	\begin{algorithm2e}
		\caption{MOGA; Assuming $B \le B^*$~\cite{FengLinearFunctionDynamicConstraintsGecco2017}, Concept from~\cite{Doerr2015FromBlackBox}.}
		\label{dynevalg:MOGA}
		\SetAlgoSkip{tinyskip}
		$P \gets$ $\{x\}$, $x$ an initial solution\;
		\While {stopping criterion not met}{
			
			\nonl~\\																% this blank line (not numbered)
			\tcc{Mutation phase.}													% this is a comment (not numbered)
			Choose $x \in P$ uniformly at random\;
			Choose $\ell$ according to $\operatorname{Bin}(n,p)$\;
			\For {$i = 1$ to $\lambda$}{
				$x^{(i)} \leftarrow \mutate_{\ell}(x)$\;
			}
			$V = \{ x^{(i)} \mid x^{(i)}$ is valid$ \}$\;
			\If {$V \neq \emptyset$}{Choose $x' \in V$ uniformly at random\;}
			\lElse{$x' \gets x$}
			
			\nonl~\\
			\tcc{Crossover phase.}
			\For {$i = 1$ to $\lambda$}{
				$y^{(i)} \leftarrow \cross_c(x,x')$\;
			}
			$M = \{ y^{(i)} \mid y^{(i)}$ is $\dominates_{\mathrm{MOEA}}$-maximal $ \wedge \, |y^{(i)}|_1 = |x|_1 +1\}$\;
			\If{$M= \{y\}$}{$y' \gets y$\;}
			\lElse{$y' \gets x$}
			
			\nonl~\\
			\tcc{Selection phase.}
			\If {$(B^* \ge |y'|_1 \ge B) \wedge (\nexists \nwspace w \in P \colon w \dominates_{\textup{MOEA}} y')$}{
				$P \gets (S \cup \{y'\}) \setminus \{z \in S \mid y' \sdominates_{\textup{MOEA}} z \}$\;
			}
		}
	\end{algorithm2e}

	Shi \etal~\cite{FengLinearFunctionDynamicConstraintsGecco2017} have also introduced a multi-objective variant of the $(1+(\lambda+\lambda))$~GA~\cite{Doerr2015FromBlackBox}, which is the fourth algorithm that they have analyzed for \om with a dynamically changing uniform constraint. In this algorithm, the same notion of dominance as MOEA is used, and the population size can grow to $D+1$. Having a solution $x$, at each iteration $\lambda$ offspring are generated by the mutation operator, which flips $l=\text{Bin}(n, p)$ random bits of $x$, where $p$ is the mutation probability. The offspring that have a 0 flipped to 1 (a 1 flipped to 0) are considered to be valid for $B^*>B$ (for $B^*<B$). One of the valid offspring (if exists), $x'$, is then used in the crossover phase, in which it is recombined with the parent solution $\lambda$ times. For a crossover probability $c$, the crossover operator creates a bit-string $y= y_1y_2\cdots y_n$, where each bit $y_i, 1 \leq i\leq n$ is chosen to be $x_i$ with probability $c$, and $x_i'$ otherwise. 
The algorithm selects the best solution $y$ with Hamming weight one larger than the Hamming weight of $x$.
The solution $y$ is added to the population if it meets the cardinality constraint and is not dominated by any other solution in the population.
	
	It is proved that this algorithm solves the \om problem with a dynamically changing uniform constraint in expected time 
$$O\left(\min\left\{\sqrt{n} D^{\frac{3}{2}}, D^2 \sqrt{\frac{n}{n-B^*}}\right\}\right)$$
 if $p=\frac{\lambda}{n}$, $c=\frac{1}{\lambda}$, $\lambda=\sqrt{n/(n-|x|_1)}$ for $B^*>B$, and in expected time 
$$O\left(\min\left\{\sqrt{n} D^{\frac{3}{2}}, D^2 \sqrt{\frac{n}{B^*}}\right\}\right)$$
 if $\lambda=\sqrt{n/|x|_1}$ for $B^*<B$~\cite{Feng2018Algorithmica}. The key argument behind these results is to show a constant probability of producing a valid offspring in the mutation phase, and then show a constant probability of generating a solution $y$ in the crossover phase that is the same as $x$ except for one bit, which is flipped from 0 to 1 for $B^*>B$ and from $1$ to $0$ for $B^*<B$.

	\subsection{Linear Pseudo-Boolean Functions Under Dynamic Uniform Constraints}
	The classical \oea and three multi-objective evolutionary algorithms have been investigated in~\cite{FengLinearFunctionDynamicConstraintsGecco2017} for re-optimizing linear functions under dynamic uniform constraints. The general class of linear constraints on linear problems leads to exponential optimization times for many evolutionary algorithms~\cite{Tobias2017UniformConstraints,Zhou2007ConstrainedOptimization}. Shi \etal~\cite{FengLinearFunctionDynamicConstraintsGecco2017} considered the dynamic setting  given in Section~\ref{dynevsec:prel:DynamicConstraints} and analyze the expected re-optimization time of the investigated evolutionary algorithms.
	This section includes the results that they have obtained, in addition to the proof ideas of their work. %The complete formal proofs can be found in the original paper.

	The algorithms that are investigated in their work, are presented in Algorithms~\ref{dynevalg:oneplusone} to~\ref{dynevalg:MOGA} of Section~\ref{dynevsec:OneMax} and the results are summarized in Table~\ref{dynevtable:LinearFunctionsResults}. The \oea (Algorithm~\ref{dynevalg:oneplusone}) uses the following fitness function which has been introduced by Friedrich \etal~\cite{Tobias2017UniformConstraints} (similar to the fitness function for \om in Section~\ref{dynevsec:OneMax}): 
	$$f_{(1+1)}(x)= f(x) - (n \nwspace w_{\max}+1) \cdot \max \nwspace \{0, |x|_1 - B^*\}$$
	Here, $w_{\max}=\max_{i=1}^n w_i$ denotes the maximum weight, and the large penalty for constraint violations guides the search towards the feasible region. 

	Shi \etal~\cite{FengLinearFunctionDynamicConstraintsGecco2017} have investigated this setting similar to the analysis of \om under dynamic uniform constraints (Section~\ref{dynevsec:OneMax}). The main difference is that for a non-optimal solution with $B^*$ $1$-bits, an improvement is not possible by flipping a single bit. A $2$-bit flip that flips a $1$ and a $0$ may be required, resulting in an expected re-optimization time of $\Or \! \left(n^2 \log (B^* \nwspace w_{\max}) \!\right)$.
	
	The second investigated algorithm, MOEA, uses the fitness function $f_{\text{MOEA}}$ and the notion of  dominance defined in Section~\ref{dynevsec:OneMax}. Unlike the re-optimization time of this algorithm for the \om problem, whose upper bound is worse than the upper bound of \oea; for the general linear functions the upper bounds obtained for MOEA are smaller than the ones obtained for \oea. The reason is that the algorithm is allowed to keep one individual for each Hamming weight between the two bounds in the population. This avoids the necessity for a 2-bit flip. To reach a solution that is optimal for cardinality $A+1$, the algorithm can use the individual that is optimal for cardinality $A$ and flip the $0$-bit whose weight is maximal. This happens in an expected number of at most $en(D+1)$ iterations, where $D=|B^*-B|$. As there are $D+1$ different cardinality values between the two bounds, the expected time to reach the optimal solution with cardinality $B^*$ is $O(nD^2)$.
	
	MOEA-S (Algorithm~\ref{dynevalg:MOEA-S}) has also been analyzed for linear functions with a dynamically changing uniform constraint. It uses single bit-flips and the population includes at most $2$ solutions: one with Hamming weight at most $B^*-1$ and one with Hamming weight $B^*$. With this setting, long waiting times for selecting a certain individual of the population are avoided. The algorithm starts with one solution in the population. It has been shown that in time $O(n \log D)$ the population consists of one solution with Hamming weight $B^*-1$ and one with Hamming weight $B^*$. Then the authors use a potential function to measure the difference of the current population to an optimal solution with Hamming weight $B^*$. The potential is given by the number of $0$-bits in the two solutions that need to be set to $1$ in order to obtain an optimal solution.
Using multiplicative drift analysis with respect to the potential function, they have proved that the expected re-optimization time of the algorithm is $O(n \log D)$.

	The fourth algorithm that is analyzed in~\cite{FengLinearFunctionDynamicConstraintsGecco2017} is MOGA (Algorithm~\ref{dynevalg:MOGA} of Section~\ref{dynevsec:OneMax}). The authors have shown that, choosing an optimal solution of Hamming weight $A< B^*$ for reproduction, an optimal solution for Hamming weight $A+1$ is produced with probability $\Omega(n^{-1/2})$ in the next generation, if $p=\frac{\lambda}{n}$, $c=\frac{1}{\lambda}$ and $\lambda=\sqrt{n}$. Since there are $D+1$ different Hamming weights to consider, and each iteration of the algorithm constructs $O(\lambda)= O(\sqrt{n})$ solutions, the expected re-optimization time is upper bounded by $O(nD^2)$.

	\begin{table*}
		\begin{center}
			\renewcommand{\arraystretch}{1.8}
			\begin{tabular}{@{}lccccc@{}}
				\toprule
				\textbf{\oea} & \textbf{MOEA} & \textbf{MOEA-S} & \textbf{MOGA} &\\
				\midrule
				$\Or(n^2 \log \nwspace (B^* \nwspace w_{\max}))$ &
				$\Or(n D^2)$ &
				$\Or(n \log D)$  &
				$\Or(n D^2)$ \\
				\bottomrule\\
			\end{tabular}
		\end{center}
		\caption{ Upper bounds on the expected re-optimization time of evolutionary algorithms on linear functions with a dynamic uniform constraint.}
		\label{dynevtable:LinearFunctionsResults}
	\end{table*}

\subsection{The Vertex Cover Problem}
The common representation for solving the vertex cover problem by means of evolutionary algorithms is the node-based representation~\cite{DBLP:journals/ec/FriedrichHHNW10,DBLP:journals/tec/OlivetoHY09,KratschFrank2013VertexCover, PPSN2016WeightedVCP}. A different representation, the edge-based representation, has been suggested and analyzed in~\cite{DBLP:conf/foga/JansenOZ13} for the static vertex cover problem. 
In this representation a search point is a bit-string $x\in \{0,1\}^m$,  where $m$ denotes the number of edges in the given graph $G=(V,E)$. For a given search point $x$, $E(x)=\{e_i\in E \mid x_i=1\}$ is the set of chosen edges. The cover set induced by $x$, denoted by $V_C(x)$, is the set of all nodes that are adjacent to at least one edge in $E(x)$.

Three variants of \RLS and \oea have been investigated. This includes one node-based approach and two edge-based approaches. The node-based approach and one of the edge-based approaches use a standard fitness function, 

$$f(s) = |V_C(s)| + (|V|+1)\cdot |\{e\in E| e\cap V_C(s) = \emptyset\}|,$$ 

in which each uncovered edge obtains a large penalty of $|V|+1$. In~\cite{DBLP:conf/foga/JansenOZ13}, an exponential lower bound for finding a 2-approximate solution for the static vertex cover problem with these two approaches using the fitness function $f$ has been shown. Furthermore, considering the dynamic vertex cover problem, Pourhassan \etal~\cite{UsDVCGecco2015} have proved that there exist classes of instances of bipartite graphs where dynamic changes on the graph lead to a bad approximation behavior.

The third variant of an evolutionary algorithm that Jansen \etal~\cite{DBLP:conf/foga/JansenOZ13} have investigated, is an edge-based approach with a specific fitness function. The fitness function $f_e$  has a very large penalty for common nodes among selected edges. It is defined as

\begin{eqnarray*}
f_e(s) =&  & |V_C(s)| + (|V|+1)\cdot |\{e\in E \mid e\cap V_C(s) = \emptyset\}|\\ 
& +& (|V|+1)\cdot(m+1)\cdot
 |\{(e,e^\prime)\in E(s)\times E(s) \mid e\ne e^\prime, e\cap e^\prime \neq \emptyset\}|\label{dyneveq:fit_func}.
\end{eqnarray*}

This fitness function guides the search towards a matching, and afterwards to a maximal matching.  In other words, whenever the algorithms find a matching, then they do not accept a solution that is not a matching, and whenever they find a matching that induces a node set with $k$ uncovered edges, then they do not accept a solution with $k'>k$ uncovered edges.
It is well known that taking all the nodes belonging to the edges of a maximal matching for a given graph results in a $2$-approximate for the vertex cover problem.

The variant of \RLS and \oea work with  the edge-based representation and the fitness function $f_e$. Note that search points are bit-strings of size $m$, and the probability of flipping each bit in \oea is $1/m$.
 Jansen \etal~\cite{DBLP:conf/foga/JansenOZ13} have proved that \RLS and \oea with the edge-based approach find a maximal matching which induces a $2$-approximate solution for the vertex cover problem in expected time $O(m\log m)$, where $m$ is the number of edges. 
 
 The behavior of \RLS and \oea with this edge-based approach has been investigated on the dynamic vertex cover problem (see Section~\ref{dynevsec:prel:DynamicVertexCover1})  in~\cite{UsDVCGecco2015}% and~\cite{DynamicVertexCoverFOCI2017}
 .  It is proved in~\cite{UsDVCGecco2015} that starting from a 2-approximate solution for a current instance of the problem, in expected time $O(m)$ \RLS finds a $2$-approximate solution after a dynamic change of adding or deleting an edge. The authors of that paper have investigated the situation for adding an edge and removing an edge separately. For adding an edge, they have shown that the new edge is either already covered and the maximal matching stays a maximal matching, or it is not covered by the current edge set and the current edge set is a matching that induces a solution with one (only the new edge) uncovered edge. Since the number of uncovered edges does not grow in this approach and the algorithm selects the only uncovered edge with probability $1/m$, a maximal matching is found in expected $m$ steps. This argument also holds for \oea, but the probability of selecting the uncovered edge and having no other mutations with this algorithm is at least $1/(em)$. Therefore, the expected re-optimization time for \oea after a dynamic addition is also $O(m)$.

When an edge is deleted from the graph, if it had been selected in the solution, a number of edges can be uncovered in the new situation. All these uncovered edges had been covered by the two nodes of the removed edge, and can be partitioned into two sets $U_1$ and $U_2$, such that all edges of each set share a node. Therefore, if the algorithm selects one edge from each set (if any exist), the induced node set becomes a vertex cover again. It will again be a maximal matching and therefore a 2-approximate solution. On the other hand, no other one-bit flips in this situation can be accepted, because they either increase the number of uncovered edges, or make the solution become a non-matching. With \RLS, in which only one-bit flips are possible, the probabilities of selecting one edge from $U_1$ and $U_2$ at each step are $\frac{|U_1|}{m}$ and $\frac{|U_2|}{m}$, respectively. Therefore, in expected time $O(m)$ one edge from each set is selected by the algorithm.

The analysis for \oea dealing with a dynamic deletion is more complicated, because multiple-bit flips can happen. In other words, it is possible to deselect an edge and uncover some edges at the same step where an edge from $U_1$ or $U_2$ is being selected to cover some other edges. An upper bound of $O(m \log m)$ is shown in~\cite{UsDVCGecco2015} for the expected re-optimization time for \oea after a dynamic deletion, which is the same as the expected time to find a 2-approximate solution with that algorithm, starting from an arbitrary solution.

	\subsection{Makespan Scheduling}
	Makespan scheduling is another problem which has been considered in a dynamic setting \cite{neumann2015runtime}.
	%In this problem, $n$ jobs and their processing times $p_i>0$, $1\leq i \leq n$ are given and the goal is to assign all job to two machines $M_1$ and $M_2$ such that the makespan is minimized, \ie to minimize the time that the busier machine takes to finish all assigned jobs. A solution is then represented by a vector $x\in \{0,1\}^n$ which means that job $i$ is assigned to machine $M_1$ if $x_i=0$ and it is assigned to $M_2$ if $x_i = 1$ for $1\leq i\leq n$. With this representation, the makespan of a given solution will be calculated as : $$f(x)=\max\left\{\sum_{i=1}^{n}{p_i(1-x_i)},\sum_{i=1}^{n}{p_ix_i}\right\}$$ and the goal is to minimize $f$.
	It is assumed that the processing time of job $i$ for $1\leq i\leq n$, is $p_i\in [L,U]$, where $L$ and $U$ are lower and upper bounds on the processing time of jobs respectively. In addition, the ratio between the upper bound and the lower bound is denoted by $R=U/L$. The runtime performance of \oea and \RLS is studied in terms of finding a solution with a good discrepancy and it is assumed that there is no stopping criteria for the algorithms except achieving such a solution. The discrepancy $d(x)$ of a solution $x$ is defined as
	$$d(x)=\left|\left(\sum_{i=1}^{n}{p_i(1-x_i)}\right)-\left(\sum_{i=1}^{n}{p_ix_i}\right)\right|.$$
	Note that a solution that has a smaller discrepancy also has a smaller makespan.
	Moreover, the proofs benefit from an important observation about the fuller machine (the machine which is loaded heavier and determines the makespan). The observation is on the minimum number of jobs of the fuller machine in terms of $U$ and $L$. :
	\begin{itemize}
		\item Every solution has at least $\lceil(P/2)/U\rceil \geq \lceil(nL/2)/U\rceil = \lceil(n/2)(L/U)\rceil=  \lceil(n/2)R^{-1})\rceil$ jobs on the fuller machine, where $P=\sum_{i=1}^{n}{p_i}$
	\end{itemize}   
	
	Two dynamic settings are studied for this problem. The first one is called  the \emph{adversary model} in which a strong adversary is allowed to choose at most one arbitrary job $i$ in each iteration and change its processing time to a new $p_i\in[L,U]$. It is proven that, independently of initial solution and the number of changes made by the adversary, \RLS obtains a solution with discrepancy at most $U$ in expected time of $O(n\min\{\log n,\log R\})$. In the case of \RLS, the number of jobs on the fuller machine increases only when the fuller machine is switched. Otherwise it increases the makespan and will not be accepted by the algorithm. This fact is the base of the proof. It is proved that if the fuller machine switches (either by an \RLS step, which moves a single job between machines, or by a  change that the adversary makes), then a solution with discrepancy at most $U$ has been found in a step before and after the switch.
	
	The proof for \oea is not as straightforward as for \RLS, since \oea may switch multiple jobs between the machines in one mutation step. However, it is shown that the number of incorrect jobs on the fuller machine, which should be placed on the other machine to decrease the makespan, has a drift towards zero. Using this argument, it is shown that \oea will find a solution with discrepancy at most $U$ in expected time $O(n^{3/2})$. Whether the better upper bounds such as $O(n\log{n})$ are possible is still an open problem.
	
	In the same dynamic setting, recovering a discrepancy of at most $U$ is also studied for both \RLS and \oea algorithms. It is assumed that the algorithm has already achieved or has been initialized by a solution with the discrepancy of at most $U$ and the processing time of a job changes afterwards. By applying the multiplicative drift theorem on the changes of the discrepancy and using the fact that the discrepancy will change by at most $U-L$, it is proven that \oea and \RLS recover a solution with discrepancy of at most $U$ in expected  time of $O(\min\{R,n\})$.
	
	The makespan scheduling problem has also been studied in another dynamic setting. In this model which is called the \emph{random model}, it is assumed that all job sizes are in $\{1,\dots,n\}$. At each dynamic change, the adversary chooses one job $i$ and its value will change from $p_i$ to $p_i-1$ or $p_i+1$ each with probability of $1/2$. The only exceptions are $p_i=n$ and $p_i=1$ for which it changes to $p_i=n-1$ and $p_i=2$, respectively.  Overall, this setting has less adversarial power than the \emph{adversary model} due to the randomness and changes by only $1$ involved.. 
	
	Let the random variable $X_i$ denotes the random processing time of job $i$ at each point of time. The following lemma proves that no large gap exists in the value of processing times which are randomly chosen for jobs.
	
	\begin{lemma}[Lemma 4 in \cite{neumann2015runtime}]
		Let $\phi(i) := |\{X_j \mid X_j=i \land j\in\{1,\dots,n\}\}|$ where $i\in\{1,\dots,n\}$, be the frequency of jobs of size $i$. Let $$G:= \max\{l \mid \exists i : \phi(i)=\phi(i+1)=\dots=\phi(i+l)=0\}$$ is the maximum gap size, \ie maximum number of intervals with zero frequency everywhere. Then, for some constant $c>0$, $$\Pr(G\geq l)\le n2^{-cl}\text{.}$$
	\end{lemma}

 This lemma states that, for any constant $c>0$ and gap size $G \geq c^\prime \log{n}$ with a sufficiently large $c^\prime$,  there is no gap of size $G$ with probability at least $1-n\cdot n^{-c-1} = 1-n^{-c}$. This probability is counted as a high probability in this study. 
	
	When the discrepancy is larger than G, it is proven that it decreases by at least one if two jobs swap between the fuller and the emptier machines. Furthermore, the maximum possible discrepancy for an initial solution is $n^2$ when all the jobs have the processing time of $n$ and are placed on one machine. Finally, it is proven that regardless of the initial solution, \oea obtains with high probability a discrepancy of at most $O(\log{n})$ after a one-time change in time $O(n^4\log{n})$. %\merk{check this!!} done
	
	The previous result considered the worst-case initial solution. However, it is proven that if the initial solution is generated randomly, then its expected discrepancy is $\Theta(n\sqrt{n})$ and it is $O(n\sqrt{n}\log{n})$ with high probability. Thus, with a random initial solution, \oea obtains a discrepancy of $O(\log{n})$ after a one-time change in time $O(n^{3.5}\log^2{n})$  with high probability.

	The two results on \oea and in the random model are for a one-time change. In extreme case, however, the processing time of a job may increase or decrease by one in each step which makes it hard to obtain a discrepancy of $O(\log{n})$, unlike the other results in this setting. Although, by using the results in the adversary model and considering that $R = U = n$, it is possible to find a solution with discrepancy of at most $n$. 
In the final theorem of this study, it is proven that independently of the initial solution and the number of changes,  \oea and \RLS obtain a solution with discrepancy of at most $n$ in expected time $O(n^{3/2})$ and $O(n\log{n})$, respectively. In addition, it is shown that the expected ratio between the discrepancy and the makespan is $6/n$. This is done by considering that a solution of discrepancy at most $n$ is obtained together with a lower bound on the makespan.
	The expected sum of all processing times is $n(n+1)/2$ and it is at least $n^2/3 +n$ with the probability of $1-2^{-\Omega(n)}$. Hence, the expected makespan is at least $n^2/6+n/2$. Furthermore, if the sum of processing times is less than $n^2/3 +n$, then the ratio would be at least $n/n$ since the processing times are at least one. Hence, if $n$ is not too small the ratio is bounded from above by
$$\frac{6}{n}-\frac{3}{n}+2^{-\Omega(n)}\leq\frac{6}{n}\text{.}$$

\subsection{The MAZE Problem}
\label{sec:eamaze}
The dynamic pseudo-boolean function \textsc{Maze} proposed in~\cite{DBLP:conf/ppsn/KotzingM12}, consists of $n + 1$ phases of $t_0 = kn^3 \log n$ iterations. During the first phase, the function is equivalent to \om. In the next $n$ phases, all bit-strings except two, still have the value equivalent to \om. The two different bit-strings, for each phase $p$ are $0^p1^{n-p}$ and $0^{p-1}1^{n-p+1}$ which have fitness values with an oscillating pattern: for two iterations out of three, these two bit-strings are assigned values $n+2$ and $n+1$, respectively, and at the third iteration, this assignment is reversed. Note that during the last phase, \textsc{Maze} behaves similar to \textsc{Trap}. The formal definition of \textsc{Maze} follows:
\begin{eqnarray*}
\textsc{Maze}(x,t) &=& 
\begin{cases}
  n+2	 \ \qquad \qquad \text{if} \  t> (n+1)\cdot t_0 \  \wedge x=0^n\\    
  n+2 \ \qquad \qquad \text{if} \ t>t_0 \  \wedge x=\text{OPT}(t)\\
  n+1 \ \qquad \qquad \text{if} \ t>t_0 \  \wedge x=\text{ALT}(t)\\
  \om(x) \quad  \text{otherwise}
\end{cases}\\
\text{OPT}(t) &=&
\begin{cases}
  \text{OPT}_{\lfloor t/t_0 \rfloor} \qquad \text{if} \  t\neq 0 \mod 3\\
  \text{ALT}_{\lfloor t/t_0 \rfloor} \qquad \text{otherwise}
\end{cases}\\
\text{ALT}(t) &=&
\begin{cases}
  \text{ALT}_{\lfloor t/t_0 \rfloor} \qquad \text{if} \  t\neq 0 \mod 3\\
  \text{OPT}_{\lfloor t/t_0 \rfloor} \qquad \text{otherwise}
\end{cases}\\
\text{OPT}_p &=& 0^p1^{n-p} \ \text{for}\  p \leq n \\
\text{ALT}_p &=& 0^{p-1}1^{n-p+1} \ \text{for}\  p \leq n 
\end{eqnarray*}

While it was shown in~\cite{DBLP:conf/ppsn/KotzingM12} that a \oea loses track of the optimum for this problem and requires with high probability an exponential amount of time to find the optimum, Lissovoi and Witt~\cite{Lissovoi2016} have proved that the optimum of the \textsc{Maze} function extended to finite alphabets, can be tracked by a \mpoea when the parent population size $\mu$ is chosen appropriately and a genotype diversity mechanism is used.

In another work~\cite{Lissovoi2017}, the behavior of parallel evolutionary algorithms is studied on the \textsc{Maze} problem. In their analysis, it is proved that both the number of independent sub-populations (or islands), $\lambda$, and the length of the migration intervals, $\tau$, influence the results. When $\tau$ is small, particularly for $\tau=1$, migration occurs too often, and the algorithm behaves similar to \oplea and fails to track the \textsc{Maze} efficiently, for $\lambda= O(n^{1-\epsilon})$, where $\epsilon$ is an arbitrary small positive constant. But with a proper choice of $\tau$, more precisely $\tau=t_0$, where $t_0$ is the number of iterations in each phase in the definition of the \textsc{Maze} problem, and a choice of $\lambda= \Theta(\log n)$, the algorithm is able to track the optimum of \textsc{Maze} efficiently.

The analysis of \mpoea and parallel evolutionary algorithms on the \textsc{Maze} problem shows that both these algorithms have limitations for tracking the optimum. \mpoea not only exploits the small number of individuals among which the optimum is oscillated, but also requires genotype diversity and a proper choice of $\mu$. On the other hand, the obtained positive results of parallel evolutionary algorithms on the \textsc{Maze} problem depend on a careful choice of migration frequency. But on the plus side, with parallel evolutionary algorithms, the problem can be extended to a finite-alphabet version.

	\section{Analysis of Evolutionary Algorithms on Stochastic Problems}
	\label{dynevsec:noise}
	The performance of \oea in noisy environment has been considered by Droste for the first time~\cite{droste2004analysis}. He proved that for the prior noise which flips a randomly chosen bit with probability $p$, \oea is able to deal with the noisy \om in polynomial time if and only if $p=O(\log(n)/n)$. Otherwise, the optimization time for $p=\omega(\log(n)/n)$ is super polynomial. 
	
	Recently, Gie{\ss}en and K{\"o}tzing~\cite{giessen2016robustness} considered \oea together with population based evolutionary algorithms in different noisy environments.
They also reproved the results of Droste with new basic theorems and studied other prior and posterior noises against \oea on \om problem. 
	
	The new prior noise models that have been analyzed recently are
	\begin{enumerate}
		\item noise which flips each bit independently with the probability of $p$
		\item noise which assign $0$ to each bit independently with the probability of $p$.
	\end{enumerate}
	\oea is able to find the optimal solution on \om or both noise models in polynomial time only if $p= O(\log(n)/n^2)$ and the optimization time grows super polynomial if $p=\omega(\log(n)/n^2)$. 
	
%\textbf{	What does this mean? if $D$ is exponentially distributed with parameter 1}
	
	The study has also covered the impact of two posterior noises on the performance of \oea. It states that \om problem under an additive posterior noise from the random variable $D$ with variance of $\sigma^2$ is tractable in polynomial time with \oea if $\sigma^2=O(\log(n)/n)$. In other case, if $D$ is exponentially distributed with parameter 1, \oea is able to find the optimum only in super polynomial time.  Furthermore,  analyzing the behavior of \oea on \om under posterior noise coming from a random variable with the Gaussian distribution $D\sim \mathcal{N}(0,\sigma^2)$, shows that it is able to deal with this noise in polynomial time if $\sigma^2\leq1/(4\log n)$. But if $\sigma^2 \geq c/(4\log n)$ for any $c>1$ then \oea finds the optimal solution of the noisy \om in super polynomial time.

	In addition to the runtime analysis, there are other measurements to rank the behavior of algorithms against the noisy problems. The concept of regret, for example, considers the progress of algorithms in approximating the noisy optimal solutions. Different definitions of regrets and how they describe the performance of algorithms have been discussed in~\cite{astete2016analysis}. As a brief introduction, to measure the approximated solution achieved by an algorithm, Simple Regret (SR$_n$) uses the solutions of $n$th iteration while Approximate Simple Regret (ASR$_n$) considers the closest solution to the optimum which has been produced until the $n$th iteration. Hence, algorithms that do not use elitism may have better performance in terms of ASR in comparison with SR measurement. 
	
	This section continues by considering the studies on the influence of using population in evolutionary algorithms for problems with noisy fitness functions.  After this, some results on the performance of a population based evolutionary algorithm against different noises is presented. Finally, we introduce an approach that modifies the algorithms by increasing the number of fitness evaluation to deal with the noise.

	\subsection{Influence of The Population Size}\label{dynevsubsec:poponEA}
	%Mostly Timo’s paper and some works by Tobias
	
In this section, we consider the impact of using populations in evolutionary algorithms against noisy problems. Gie{\ss}en and K{\"o}tzing~\cite{giessen2016robustness} studied this matter by considering \mpoea and \oplea on the noisy \om. Considering a noisy function $f$, and let $(X_k)_{k\leq n}$ be a random variable taking on the value of the noisy function $f$ for a solution with exactly $k$ ones. It is also assumed that  $\forall j : 0<j<k<n $ we have $\Pr(X_k<X_{k+1})\leq \Pr(X_j<X_{k+1})$. This means that when solutions are close, it is more likely to observe a confusion caused by the noise. The analysis of the performance of \mpoea on \om with prior noise is based on the following theorem.
	
	\begin{theorem}[Theorem 12 in \cite{giessen2016robustness}]\label{dynevthe:mu+1theorem}
		Let $\mu$ be given and suppose for each $k\leq n$, $X_k\in [k-1,k+1]$. For each $k < n$, let $A_k$ be the event that drawing $\mu$ independent copies of $X_k$ and one copy of $X_{k+1}$ and the sorting with breaking ties uniformly, the value of $X_{k+1}$ does not come out least. If there is a positive constant $c\leq 1/15$ such that $$\forall k,n/4<k<n: \Pr(A_k)\geq 1-c\frac{n-k}{n\mu}\text{,}$$then \mpoea optimizes $f$ in an expected number of $O(\mu n\log{n})$ iterations.  
	\end{theorem}
	
	The proof of this theorem is based on the definitions of two events to show that there is a positive drift on the number of ones in the best solution which has $k$ ones in the current step of the algorithm. The first event, $E_0$, is the event that the new solution has at least a bit with value one more than the current solution and it is not dominated by any other solutions, even when considering the noise. The other event, $E_1$, is the situation that the new solution has less ones than the current best solution, the current best solution is unique and it is ignored because of the noisy function. For this case to happen, the best solution with $k$ ones must be evaluated to have $k-1$ ones, all other solutions must have at least $k-2$ ones and be evaluated to have at least $k-1$ ones because of the noisy function. After this event, the number of ones of the best solution decreases at most by $2$. Considering the probability of each event, the drift on the number of ones in the best solution is at least
$$\frac{n-k}{e\mu n}-3\frac{c(n-k)}{n\mu}-2c/(n\mu).$$ 
Finally, since $c\leq 1/15$ and using multiplicative drift analysis the theorem is proven.
	
	The previous theorem is used to prove a corollary for the performance of \mpoea on noisy \om with the prior noise, \ie the noise which flips a bit uniformly at random with probability $p$. It is proven that if $\mu \geq 12\log{(15n)}/p$ then \mpoea finds the optimum of \om in expected number of $O(\mu n\log{n})$ iterations. To be more specific, $\mu=24\log{(15n)}$ is adequate to achieve such expected time for $p=1/2$. 
	
	Gie{\ss}en and K{\"o}tzing also considered the performance of \oplea as another population based evolutionary algorithm on the noisy \om problem. In \oplea there exists an offspring population. The algorithm produces an offspring with size  $\lambda$ by mutating the current best solution $\lambda$ times. Then, it chooses the best solution among the offspring and the parent as the next best solution. They prove a main theorem to achieve the results on this matter. The theorem is as follows:
	
	\begin{theorem}[Theorem 14 in \cite{giessen2016robustness}]\label{dynevthe:onepluslambda}
		Let $\lambda\ge24\log{n}$ and, for each $k< n$, let $Y_k$ denote the maximum over $\lambda$ observed values of $X_k$ (belonging to inferior individuals) and let $Z_k$ denote the maximum over at least $\lambda/6$ observed values of $X_k$ (belonging to better individuals). Suppose there is $q < 1$ such that
		\begin{align}\label{dynevequ:condition1}
		\forall k<n: \Pr(Y_k<X_{k+1})\ge q\text{,}
		\end{align}
		and
		\begin{align}\label{dynevequ:condition2}
		\forall k<n : \Pr(Y_{k-1}<Z_k)\geq 1-\frac{q}{5}\frac{l\lambda}{en+l\lambda}\text{.}
		\end{align}
		Then \oplea optimizes $f$ in $O((\frac{n\log{n}}{\lambda}+n)/q)$ iterations and needs $O((n\log{n}+n\lambda)/q)$ fitness evaluations.  
	\end{theorem}
	In the theorem, $l$ is the number of zeros of the current best solution. To prove it, similar to Theorem \ref{dynevthe:mu+1theorem}, it is shown that the drift on the number of ones is positive and equal to $(q-\frac{4q}{5})\frac{l\lambda}{en+l\lambda}$. Furthermore, this theorem gives the sufficient conditions (Equations \ref{dynevequ:condition1} and \ref{dynevequ:condition2}) on noises to demonstrate whether they are tractable with \oplea in a guaranteed expected number of iterations.
	
	As a corollary, for the prior noise which flips a bit uniformly at random with probability $p$, it is proven that \oplea with $\lambda\geq \max\{12/p,24\}n\log{n}$ optimizes \om in expected time $O((\frac{n^2\log{n}}{\lambda}+n^2)/p)$. Let $q=p/n$. To show that Equation \ref{dynevequ:condition1} holds, it is enough to consider the event that the solution with $k$ ones (current best solution) is evaluated to have more ones. The complement of this event is a set of events that either there is no noisy bit flip or the noise flips a zero bit, which leads to:$$\Pr(Y_k<X_{k-1})\geq 1- \left(1-p+\frac{pk}{n}\right)\geq\frac{p}{n}=q\text{.}$$ With a similar consideration about the probability of improving at least one of $\lambda/6$ of the solutions, it is observed that Equation \ref{dynevequ:condition2} also holds and the corollary is correct.
	
	The other corollary from Theorem \ref{dynevthe:onepluslambda} is about the non-positive additive posterior noise $D$ that is evaluated to $>-1$ with a non-zero probability $p$. It is proven that under this noise, for $\lambda\geq\max\{10e,\frac{-6\log{(n/p)}}{\log(1-p)}\}$, \oplea optimizes \om in time $O((n\log{n}+n\lambda)/p)$. The proof of this corollary is a bit more tricky. Since $D\leq0$, we have $Y_k\leq k$ and 
$$\Pr(Y_k\geq X_{k+1})\leq\Pr(k\geq X_{k+1}) = \Pr(D\leq-1)=1-p\text{.}$$ 

This means that $\Pr(Y_k<X_{k+1})\geq p$ which fulfills the first condition of Theorem \ref{dynevthe:onepluslambda}. To consider the second condition, a similar complement technique is used, \ie calculating the probability of $Y_{k-1}\geq Z_k$ which is the complement of $Y_{k-1}<Z_k$. To satisfy $Y_{k-1}\geq Z_k$, all of $\lambda/6$ solutions which have $k$ ones should be affected by a noise with the value less than $-1$; thus  $\lambda \geq \frac{-6\log{(n/p)}}{\log(1-p)}$ concludes that $\Pr(Y_{k-1}\geq Z_k)\leq(1-p)^{\lambda/6} \leq p/n\text{.}$ Finally, since $\lambda \geq 10e$, the second condition of  $\Pr(Y_{k-1}<Z_k)$ has been proven to be satisfied; therefore, Theorem \ref{dynevthe:onepluslambda} holds.

	\subsection{Influence of Different Noise Distributions on The Performance of Population Based EAs}
	%Mostly works by Tobias. I guess this chapter will also include different noise distributions that are analysed by Tobias
	
	Friedrich \etal~\cite{TobiasTimoNoise2017,friedrich2016graceful} have considered \mpoea and additive posterior noises with different distributions. They introduced the concept of "graceful scaling" to determine the performance of an algorithm against noises. An algorithm scales gracefully with noise if there exists a parameter setting for the algorithm such that it finds the optimum of the real fitness function when evaluating the noisy one, in a polynomial time.
	
	%\remark{the exact definition of graceful scaling will cause new definition which are not used at all, should I bring it or reword the definition?}
	
	They also introduce  a sufficient condition that \mpoea cannot deal with noise and is unable to find the real optimum through the noisy function. The condition is: if there are $\mu + 1$ different values $\{d_1,\dots,d_{\mu+1}\}$ from the random variable $D$, $Y$ is the minimum of $\{d_1,\dots,d_\mu\}$ and $$\Pr(Y>d_{\mu+1} +n)\geq\frac{1}{2(\mu+1)}\text{,}$$ then \mpoea, with a polynomially bounded $\mu$ will not evaluate the optimum with high probability.
	%\remark{how to prove this?}
	
	This theorem is used to analyze the performance of \mpoea with $\mu = \omega(1)$ and bounded from above by a polynomial, on noisy \om problem with different distributions. In this study it is proven that if the noise comes from a Gaussian distribution with $\sigma^2\geq(na)^2$, for some $a=\omega(1)$, \mpoea will not find the optimum in polynomial time with high probability.
	%\remark{the proof?}
	
	Furthermore, other noise distributions have been studied in \cite{friedrich2016graceful}. They analyzed a random variable $D$ with a distribution that decays exponentially. Here the probability density function of $D$ is as follows:
	
	\begin{align}\nonumber
	F(t)&:=\Pr(D<t)=\frac{1}{2}e^{ct} &&\text{if } t\leq0 \text{ and}\\\nonumber
	F(t)&:=1- \frac{1}{2}e^{-ct} &&\text{if } t\ge 0\text{,}
	\end{align}
	for some constant $c$ and variable $t$. The probability mass function $p$ of $D$ is obtained by taking the derivative of $F$:
	\begin{align}\nonumber
	p(t) & = F^\prime(t) = \frac{c}{2}e^{ct}&&\text{if }t\leq 0 \text{ and}\\\nonumber
	p(t) & = \frac{c}{2}e^{-ct} && \text{if } t\ge0\text{.}
	\end{align}
	Note that this is a symmetric variant of an exponential distribution. It is observed that $p$ is symmetric around $0$ which implies that the distribution of $D$ has mean $0$.
	
	The variance of $D$ is calculated as follows:
	
	\begin{align}\nonumber
	\text{Var($D$)}&= \int_{-\infty}^{+\infty}{t^2p(t)\text{d}t} =\frac{c}{2}\left(\int_{-\infty}^{0}{t^2e^{ct}\text{d}t} + \int_{0}^{\infty}{t^2e^{-ct}\text{d}t} \right)\\\nonumber
	&=c\int_{-\infty}^{0}{t^2e^{ct}\text{d}t}=\left[\frac{(2-2ct+t^2c^2)e^{ct}}{c^3}\right]^0_{-\infty}\\\nonumber
	&=\frac{2}{c^2}=:\sigma^2\text{.}
	\end{align}
	Now $F(t)$ can be rewritten in terms of $\sigma$:
	\begin{align}\nonumber
	F(t)&:=\frac{1}{2}e^{\sqrt{2}\frac{t}{\sigma}} &&\text{if } t\leq0 \text{ and}\\\nonumber
	F(t)&:=1- \frac{1}{2}e^{-\sqrt{2}\frac{t}{\sigma}} &&\text{if } t\ge 0\text{.}
	\end{align}
	
	In this setting, it is proven that if the variance is large ($\sigma^2 = \omega(n^2)$), then \mpoea will not find the optimum of  the noisy \om. The proof applies the condition that is mentioned above and bounds $\Pr(Y>d_{\mu+1}+n)$. The idea is to consider a subset of events in which $Y>d_{\mu+1}+n$ holds. To this aim, points $t_0$ and $t_1$ are defined such that $t_0<t_1$ and we have $\Pr(D<t_0)$ and $\Pr(D<t_1)$ dependent on $\mu$ in a way that  $\Pr(D<t_0)<\Pr(D<t_1)<1/2$. This definition leads to the following events:
	
	\begin{itemize}
		\item[A:] The event that $D<t_0-n$ and $t_0<Y$.
		\item[B:] The event tat $t_0-n<D<t_1-n$ and $t_1<Y$
	\end{itemize} 
	
	The fact that $\sigma^2 \geq (na)^2$ for $a=\omega(1)$, helps to find the lower bounds for the probabilities and results in $\Pr(Y>d_{\mu+1})\geq\frac{1}{2(\mu+1)}$. Hence, \mpoea will not find the optimum of noisy \om if the noise comes from an exponential distribution as defined.
	
	The other noise distributions which are studied by Friedrich \etal~\cite{friedrich2016graceful} are \emph{Truncated Distributions}. It is proven that \mpoea scales gracefully with these kind of noises which are generalization of uniform distribution.
	
	\begin{definition}[Definition 7 in \cite{friedrich2016graceful}]
		Let $D$ be a random variable. If there are $k,q\in \R$ such that $$\Pr(D>k) = 0 \text{  }\land\text{  } \Pr (D\in (k-1,k])\geq q\text{,}$$ then $D$ is called upper $q$-truncated. Analogously, $D$ is called lower $q$-truncated if there is are $k,q\in\R$ with $$\Pr(D<k) = 0 \text{  } \land\text{  }\Pr (D\in [k,k+1))\geq q\text{.}$$ 
	\end{definition}

	Using this definition, it is proven that \mpoea obtains the optimum of noisy \om with a lower $2\log(n\mu)/\mu$-truncated noise distribution in expected $O(\mu n\log{n})$ iterations. The proof uses multiplicative drift and benefits from the fact that the best solution is never removed in the first $O(\mu n\log{n})$ iterations if any other point is evaluated in the minimal bracket $[k,k+1)$. The first corollary of this result is that for an arbitrary lower $q$-truncated noise, \mpoea with $\mu\geq3^{-1}\log(nq^{-1})$ evaluates the optimum of noisy \om after expected $O(\mu n\log n)$ iterations.
	
	Finally the last corollary in this study considered a uniform distribution on $[-r,r]$, which is lower $1/2r$-truncated, as the noise function. In this manner, by using the previous results, it has been proven that \mpoea scales gracefully on \om with additive posterior noise from the uniform distribution on $[-r,r]$. 
	
	\subsection{Resampling Approach for Noisy Discrete Problems}

In the previous sections, we have considered the behavior of evolutionary algorithms for noisy problems. This section presents another approach of dealing with such problems. Here, a modified version of the algorithm, which has a known performance on the noise-free case, is investigated. Akimoto et al. have studied resampling methods to modify iterative algorithms and found upper bounds on its performance according to the proven performance of the known algorithm~\cite{akimoto2015analysis}. Their presented framework suits EAs perfectly. In a resampling method, the evaluation of the noisy fitness function for each solution is repeated $k$ times. The algorithm then takes the average of $k$ noisy values as the fitness value of the solution. Let $Opt$ and $k$-$Opt$ denote the original algorithm and the resampling modified version, respectively. The parameter $k$ can be fixed, or be adapted during the optimization process. In~\cite{akimoto2015analysis}, the authors have classified discrete optimization problems in two different categories. Either there is a known algorithm available that finds the optimal solution in expected $r(\delta)$ fitness evaluations with probability $1-\delta$, or no such an algorithm is known. In the first case, $k$ is chosen according to $r(\delta)$ and in the second case, its value is set adaptively in each iteration. 

Assume that the additive Gaussian noise with variance $\sigma^2$ is applied to the fitness function. In the pre-known runtime case, it is proven that by fixing $$k_{g}=\max\left(1,\left\lceil 32\sigma^2\left[\ln(2)-\ln\left(1-(1-\delta)^{1/r(\delta)}\right)\right] \right\rceil \right)\text{,}$$ $k_{g}$-$Opt$ finds the optimum of the noisy function with probability at least$(1-\delta)^2$ and the expected running time is $$O\left(r(\delta)\max\left(1,\sigma^2\ln\left(\frac{r(\delta)}{\delta}\right)\right)\right)\text{.}$$ 

Let $p$ denote the probability of the ratio of noisy and real fitness values of point $x$ being at least $\frac{1}{4}$. The proof first determines $p\leq 2\exp\left(-\frac{k/4^2}{2\sigma^2}\right)$. This leads to the probability of situations that the noisy fitness value is sufficiently close to the real one. 

On the other hand, suppose there is no algorithm to solve the noise-free problem. However, let there is a known algorithm $Opt^\prime$ which satisfies criterion $Opt$ with probability at least $1-\delta$ after $n$ total fitness evaluations in iteration $n$. 

Let $\beta>1$ and $$k_m=\left\lceil32\sigma^2\left(\frac{2(n+1)\ln(n+1)^\beta}{\delta}\left(\sum_{i=2}^{\infty}\frac{1}{i\ln(i)^\beta}\right)\right) \right\rceil$$

be the number of resampling of $k$-$Opt^\prime$ in iteration $m$. 
It is proven that $k$-$Opt^\prime$ satisfies $Q$  for the noisy problem with probability at least $(1-\delta)^2$ after $n$ iteration and the total number of fitness evaluations is $$\sum_{m=1}^{n} k_m = O(n\ln n).$$ 

A more general scenario (called the heavy tail scenario) is also considered. Here, there is no assumption on the noise distribution, except that the variance, $\sigma^2$, is finite with mean zero. It is proven that by fixing $$k_{h} =\max(1,\lceil16\sigma^2/(1-(1-\delta)^{r(\delta)})\rceil),$$ 

$k_{h}$-$Opt$ solves the noisy problem with probability at least $(1-\delta)^2$ and the expected runtime is $$O\left(r(\delta)\max\left(1,\sigma^2\frac{r(\delta)}{\delta}\right)\right)\text{.}$$ 

When no algorithm is known to find the optimum, then by using the definition of $Q$, $k$-$Opt^\prime$ with $$k_m=\left\lceil\frac{16\sigma^2(n+1)\ln(n+1)^\beta}{\delta}\sum_{i=2}^{\infty}\frac{1}{i\ln(i)^\beta} \right\rceil,$$
 where $\beta>1$, satisfies $Q$ for any heavy tail noisy function with probability at least $(1-\delta)^2$. 
The total number of fitness evaluation up to iteration $n$ is $$\sum_{m=1}^{n}k_m = O(n^2\ln(n)^\beta).$$

\section{Ant Colony Optimization}
\label{sec:ACO}
\begin{figure*}
\begin{center}
\begin{tikzpicture}[scale=2]
\tikzstyle{antbody}=[fill=black,draw=black];
\tikzstyle{antleg}=[black,rounded corners=0pt];
\begin{scope}[shift={(-0.55,0)},scale=0.2]
% draw legs and antennae
% first the ones on the left-hand side of the ant
\draw[antleg] (6pt+5pt,0pt) -- ++(-5pt,+7pt) -- ++(-11pt,+1pt) -- ++(-3pt,+1pt);
\draw[antleg] (6pt+8pt,0pt) -- ++(+1pt,+7pt) -- ++(-3pt,+7pt);
\draw[antleg] (6pt+10pt,0pt) -- ++(+8pt,+6pt) -- ++(+2pt,+5pt);
\draw[antleg,rounded corners=1pt] (6pt-2pt+2*9pt-1pt+6pt+4pt,0pt) -- ++(-2pt,+8pt) -- ++(+10pt,+3pt) -- ++(+4pt,+2pt);
% then mirror to draw the same on its right-hand side
\draw[antleg] (6pt+5pt,0pt) -- ++(-5pt,-7pt) -- ++(-11pt,-1pt) -- ++(-3pt,-1pt);
\draw[antleg] (6pt+8pt,0pt) -- ++(+1pt,-7pt) -- ++(-3pt,-7pt);
\draw[antleg] (6pt+10pt,0pt) -- ++(+8pt,-6pt) -- ++(+2pt,-5pt);
\draw[antleg,rounded corners=1pt] (6pt-2pt+2*9pt-1pt+6pt+4pt,0pt) -- ++(-2pt,-8pt) -- ++(+10pt,-3pt) -- ++(+4pt,-2pt);
% draw the ant's body on top
\filldraw[antbody] (6pt-2pt+9pt,0) ellipse (9pt and 2pt);
\filldraw[antbody] (0,0) ellipse (7pt and 5pt);
\filldraw[antbody] (6pt-2pt+2*9pt-1pt+6pt,0) ellipse (5pt and 4pt);
\end{scope}
\tikzstyle{node}=[black,fill=white,thick];
\tikzstyle{edge}=[black,thick,-triangle 60];
\foreach \x in {1,2,3,4,5} {
    \draw[edge,shorten >=8pt,shorten <=8pt] (\x-1,0) .. controls (\x-1+0.3,0.4) and (\x-1+0.7,0.4) .. (\x,0) node[pos=0.5,above] {$e_{\x,1}$};
    \draw[edge,shorten >=8pt,shorten <=8pt] (\x-1,0) .. controls (\x-1+0.3,-0.4) and (\x-1+0.7,-0.4) .. (\x,0) node[pos=0.5,below] {$e_{\x,0}$};
}
\foreach \x in {0,1,2,3,4,5} {
    \filldraw[node] (\x,0) circle (4pt) node {$v_{\x}$};
}
\end{tikzpicture}
\end{center}
\caption{Construction graph for pseudo-Boolean optimization with $n=5$ bits.}
\label{fig:construction-multigraph}
\end{figure*}
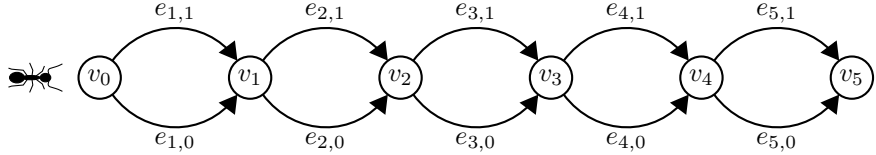

After having investigated evolutionary algorithms for dynamic and stochastic problems, we now give a summary on the results obtained in the context of ant colony optimization (ACO). ACO algorithms construct solutions for a given optimization problem by performing random walks on a so-called construction graph.
The construction graph frequently used in pseudo-Boolean optimization is shown in Figure~\ref{fig:construction-multigraph}.
 This random walk is influenced by so-called pheromone values on the edges of the graph. At each time step, the pheromone values induce a probability distribution on the underlying search space which is used to sample new solutions for the given problem.
 Pheromone values are adapted over time such that good components of solutions obtained during the optimization process are reinforced. The idea is that this reinforcement then leads to better solutions during the optimization process.
An algorithm which is frequently studied throughout theoretical investigations for pseudo-Boolean maximization is MMAS (see Algorithm~\ref{alg:MMAS}). It is a simplification of the Max-Min Ant System introduced in \cite{DBLP:journals/fgcs/StutzleH00}. The algorithm, which is given in Algorithm~\ref{alg:MMAS}, only uses one ant in each iteration. However, variants of MMAS called $\lambda$-MMAS, where in each iteration $\lambda$ ants are used, have also been studied in the literature.
 Pheromone values are chosen within the interval $[\tau_{\min}, \tau_{\max}]$ where $\tau_{\min}$ and $\tau_{\max}$ are lower and upper bounds used in MMAS. Furthermore, the update strength $\rho$ plays an important role in the runtime analysis of ACO algorithms. For MMAS, a large update strength such as $\rho=1$ often makes the considered MMAS algorithms similar to simple evolutionary algorithms such as \oea. The considered algorithms are usually analyzed with respect to the number of solutions until a given goal has been achieved. As in the case of runtime analysis of evolutionary algorithms, one is often interested in the expected number of solutions to reach the desired goal.

\renewcommand{\algorithmicloop}{\textbf{repeat forever}}
\begin{algorithm}[ht]
    \caption{\MMAS}
    \algsetup{indent=1.5em}
    \begin{algorithmic}[1]
        \STATE Set $\tau_{(u,v)} = 1/2$ for all $(u,v) \in E$.
        \STATE Construct a solution $x^*$.
        \STATE Update pheromones \wrt{} $x^*$.
        \LOOP
            \STATE Construct a solution $x$.
            \STATE \textbf{if} $f(x) \ge f(x^*)$ \textbf{then} $x^*:=x$.
            \STATE Update pheromones \wrt{} $x^*$.
        \ENDLOOP
    \end{algorithmic}
    \label{alg:MMAS}
\end{algorithm}

\subsection{Dynamic Problems}
K{\"o}tzing and Molter~\cite{DBLP:conf/ppsn/KotzingM12} compared the behavior of \oea and MMAS on the \textsc{Maze} problem.
The \textsc{Maze} problem has an oscillating behavior of different parts of the function and the authors have shown that MMAS is able to track this oscillating behavior if $\rho$ is chosen appropriately, i.e.  $\rho=\theta(1/n)$ whereas \oea looses track of optimum with probability close to $1$.

In the case of dynamic combinatorial optimization problems, dynamic single-source shortest paths problems have been investigated in \cite{LissovoiWittDynamic2015}. 
Given a destination node $t \in V$, the goal is to compute for any node $v \in V \setminus t$ a shortest paths from $v$ to $t$. The set of these single-source shortest paths can be represented as a tree with root $t$ and the path from $v$ to $t$ in that tree gives a shortest path from $v$ to $t$.
The authors have investigated different types of dynamic changes for variants of MMAS. They first investigated the MMAS and show that this algorithm can effectively deal with one time changes and build on investigations in \cite{DBLP:journals/jda/SudholtT12} for the static case. 
They show that the algorithm is able to recompute single-source shortest paths in an expected number of
$O(\ell^* / \tau_{\min} + \ell \ln(\tau_{\max}/\tau_{\min})/\rho)$ iterations after a one change has happened. 
The parameter $\ell$ denotes the maximum number of arcs in any shortest path to node $t$ in the new graph and $\ell^* = \min\{\ell, \log n\}$.~The result shows that MMAS is able to track dynamic changes if they are not too frequent.
Furthermore, they present a lower bound of $\Omega(\ell/\tau_{\min})$ in the case that $\rho=1$ holds.
Afterwards, periodic local and global changes are investigated. In the case of the investigated local changes, $\lambda$-MMAS with a small $\lambda$  is able to track periodic local changes for a specific setting. For global changes, a setting  with oscillation between two simple weight functions is introduced where an exponential number of ants would be required to make sure that an optimal solution is sampled with constant probability in each iteration.

\subsection{Stochastic Problems}
In stochastic environments, ACO algorithms have been analyzed for exemplary benchmark functions and the stochastic shortest paths problem. 

Thyssen and Sudholt~\cite{DBLP:conf/ppsn/2012-1} started the runtime analysis of ACO algorithms in stochastic environments. They investigated the single-destination shortest path (SDSP)  problem where edge weights are subject to noise. For each edge $e$ the noise model return a weight of $(1 + \eta(e, p, t))\cdot  w(e)$ instead of the exact real weight $w(e)$. This implies that the weight $w(e)$ of each edge $e$ is increased according to the noise model.
They considered a variant of MMAS for the shortest path problem introduced in \cite{DBLP:journals/jda/SudholtT12}. They start by characterizing noise model for which the algorithm can discover shortest paths efficiently. In the general setting, they examined the algorithms in terms of approximations. The results depends on how much non optimal paths differ at least from optimal ones. More precisely, they show that  
if for each vertex $v \in V$ and some $\alpha > 1$ it holds that every non-optimal path has length at least $(1 + \alpha \cdot E(\eta(opt_v))) \cdot opt_v$ then the algorithm finds an $\alpha$-approximation in time proportional to $\alpha/(\alpha-1)$ and other standard ACO parameters such as pheromone bounds and pheromone update strengths. Here $opt_v$ denotes the value of a shortest paths from $v$ to the destination $t$ and $E(\eta(opt_v))$ denotes the expected random noise on all edges of this path.
Furthermore, for independent gamma-distributions along the edges, they have shown that the algorithm may need exponential time to find a good approximation.
Doerr et al.~\cite{DBLP:conf/gecco/DoerrHK12} have extended the investigations for the stochastic  SDSP problem. They considered a slight variation of MMAS for the stochastic SDSP which always reevaluates the best so-far solution when a new solution for comparison is obtained. This allows the MMAS version to easily obtain shortest paths in the stochastic setting.

%Optimizing expected path lengths~\cite{DBLP:conf/foga/FeldmannK13}.

Friedrich et al.~\cite{DBLP:journals/ec/FriedrichKKS16} considered MMAS with a  fitness proportional pheromone update rule on linear pseudo-Boolean functions. They show that the algorithm scales gracefully with the noise, i.e. the runtime only depends linearly on the variance of the Gaussian noise. In contrast to this many of the considered noise settings are not solvable by simple evolutionary algorithms such as \oea~\cite{droste2004analysis}. This points out a clear benefit of using ant colony optimization for stochastic problems.

\section{Conclusions}

Evolutionary algorithms have been extensively used to deal with dynamic and stochastic problems. We have given an overview on recent results regarding the theoretical foundations of evolutionary algorithms for dynamic and stochastic problems in the context of rigorous runtime analysis. Various results for dynamic problems in the context of combinatorial optimization for problems such as makespan scheduling and minimum vertex cover have been summarized and the benefits of different approaches to deal with such dynamic problems have been pointed out. In the case of stochastic problems, the impact on the amount of noise and how population-based approaches are beneficial for coping with noisy environments have been summarized. 

While all these studies have greatly contributed to the understanding of the basic working principles of evolutionary algorithms and ant colony optimization in dynamic and stochastic environments, analyzing the behavior on complex problems remains highly open. Furthermore, also uncertainties often change over time and are therefore dynamic. Therefore, it would be very interesting to analyze the behavior of evolutionary algorithms for problems where uncertainties change over time.
For future research, it would also be interesting to examine environments that are both dynamic and stochastic as many real-world problems have both properties at the same time.
	\bibliographystyle{abbrv}
	\bibliography{myreferences.bib}
%	\balance
	
\end{document}